# Evaluating Large Language Models in Ophthalmology

Jason Holmes[1]*, Shuyuan Ye[2]*, Yiwei Li[4]*, Shi-Nan Wu[2,6], Zhengliang Liu[4], Jinyu Hu[2,3], Huan Zhao[5], Xi Jiang[5], Wei Liu[1], Hong Wei[2,3], Jie Zou[2,3], Tianming Liu[4#], Yi Shao[3#]

[1]Department of Radiation Oncology, Mayo Clinic

[2]Department of Ophthalmology, the First Affiliated Hospital of Nanchang University, Nanchang 330006, China;

[3]Department of Ophthalmology, Eye& ENT Hospital of Fudan University, shanghai 200030, China;

[4]School of Computing, University of Georgia;

[5]School of Life Science and Technology, University of Electronic Science and Technology of China;

[6]Eye Institute of Xiamen University, School of Medicine, Xiamen University, Xiamen, Fujian, China.

*These authors have contributed equally to this work.

[#]**Address correspondence to:**

Yi Shao (Email: freebee99@163.com; Tel:+086 791-88692520, Fax:+086 791-88692520), Department of ophthalmology, Eye& ENT Hospital of Fudan University, shanghai 200030, China

**Conflict of Interest Statement:** This was not an industry supported study. The authors report no conflicts of interest in this work.

**Fund program:** National Natural Science Foundation of China (82160195); Jiangxi Double-Thousand Plan High-Level Talent Project of Science and Technology Innovation (jxsq2023201036); Key R & D Program of Jiangxi Province (20223BBH80014)

**Ethical Statement:** All research methods were approved by the committee of the medical ethics of the First Affiliated Hospital of Nanchang University and were in accordance with the 1964 Helsinki declaration and its later amendments or comparable ethical standards. All subjects were explained the purpose, method, potential risks and signed an informed consent form

**Abstract**

**Purpose:** The performance of three different large language models (LLMS) (GPT-3.5, GPT-4, and PaLM2) in answering ophthalmology professional questions was evaluated and compared with that of three different professional populations (medical undergraduates, medical masters, and attending physicians).

**Methods:** A 100-item ophthalmology single-choice test was administered to three different LLMs (GPT-3.5, GPT-4, and PaLM2) and three different professional levels (medical undergraduates, medical masters, and attending physicians), respectively. The performance of LLM was comprehensively evaluated and compared with the human group in terms of average score, stability, and confidence.

**Results:** Each LLM outperformed undergraduates in general, with GPT-3.5 and PaLM2 being slightly below the master's level, while GPT-4 showed a level comparable to that of attending physicians. In addition, GPT-4 showed significantly higher answer stability and confidence than GPT-3.5 and PaLM2.

**Conclusion:** Our study shows that LLM represented by GPT-4 performs better in the field of ophthalmology. With further improvements, LLM will bring unexpected benefits in medical education and clinical decision making in the near future.

1. Introduction

The increase of large language model (LLM) indicates that great progress is being made in the field of natural language processing (NLP)[1]. LLM is a deep learning model trained on large-scale plain text data[2, 3]. Specifically, LLM is based on transformer architecture, and it can acquire general skills (language understanding, text output, etc.) by learning existing corpora (books, newspapers, Internet, etc.). Then, through instruction tuning and alignment tuning, its ability is further enhanced to make its behavior more in line with human values and preferences, so as to better complete tasks and generate high-quality output text. The major difference between LLM and traditional small-scale language models is that its parameter capacity is larger, containing billions or even more parameters. Such a large parameter scale not only significantly improves the performance of language models, but also exhibits some functions that small-scale language models lack, such as context learning, instructions to follow, and reasoning [4].

Currently, the latest version of ChatGPT (GPT-4) released by OpenAI is considered to be one of the most powerful LLMs[5]. The new generation of LLM represented by GPT-4 shows strong functions and has already achieved success in the general field. For example, LLM can successfully complete NLP tasks in the field of food and agriculture by pre-training the model on agriculture-related text data[6]. ChatGPT also achieved considerable scores in the US Certified Public Accountant exam and the US Bar exam[7, 8]. In addition, available data suggests that LLM has great potential in fields such as medicine[9-16] that require advanced and complex thinking. Gilson found that ChatGPT met the criteria for passing the US Medical Licensing Examination (USMLE)[17]. Moreover, ChatGPT can also help surgeons collect patient history[18], analyze medical imaging features, accurately diagnose diseases, optimize surgical plans, predict surgical results, improve surgical efficiency and safety, and strengthen postoperative management and rehabilitation[19, 20]. ChatGPT was even found to perform better in some aspects than a doctor. In an experiment evaluating ChatGPT's ability to answer patients' questions, ayers found that ChatGPT outperformed doctors in both the quality and empathy of their responses to questions[21]. However, these tests are all based on basic or scientific knowledge, and there are many existing materials, and even the answers can be easily found in the existing corpus, which makes it impossible to evaluate the performance of LLM systematically and comprehensively. However, as of now, the potential risks associated with the use of LLMs in medicine include limited and outdated medical data training, leading to inaccurate medical recommendations[21]. For instance, a study involving a LLM trained in retinal-related diseases showed an identification accuracy of only 45% when it involved information sources from patients with retinal diseases[22]. This indicates a significant gap between AI applications in ophthalmology clinical settings.

Based on the above situation, to scientifically and systematically explore and analyze the model performance, it is necessary to not only select more subtle fields with higher professional barriers, but also to ensure that the test content is not included in the training data[23].To this end, in this study, we created 100 new ophthalmology single-choice questions for testing, with the aim of evaluating the performance of three different LLMs (GPT-3.5[24], GPT-4[5], and PaLM2[25]) in answering ophthalmic questions, and comparing the results with three different levels of professional populations (medical undergraduates, medical masters, and attending physicians) to match the professional populations corresponding to different models. In addition, we also discussed the stability and self-confidence of different models in the test, and will continue to explore the reliability of ChatGPT-4 for medical education and clinical decision making.

## 2. Methods

The 100 ophthalmic single-choice questions used for testing in this study were drafted by experienced ophthalmologists. The examination questions are listed in Appendix Section A.

For testing, 100 ophthalmology single-choice questions were entered into each LLM separately. Each test started with a new thread or reset initialization prompt. Subsequently, the LLM was prompted in batches of 20 questions until the end of the test. For each test, it was explained to the LLM that absolutely correct answers must be returned. If the LLM could not handle 20 questions at once, the number of batch-processing questions was changed to 10. If the LLM did not return answers to all questions in that batch, the unanswered questions were included in the next batch of tests, in addition to all questions in the next batch.

The LLM test results were compared with the results of three professional populations (medical undergraduates, medical masters, and attending physicians). Nine undergraduates of clinical medicine and six masters of ophthalmology at Nanchang University were randomly selected as the undergraduate group and the masters group, and three attending doctors of ophthalmology in the First Affiliated Hospital of Nanchang University were randomly selected as the attending doctor group. Each candidate took a 3-hour, closed-book exam. Mean scores, response stability, and answer confidence were assessed when comparing LLM and human performance.

To quantify accuracy, we characterized the score for each LLM by calculating the mean score for each trial. For the human test group, each group score was expressed as the group mean.

To quantify the response stability across LLMs, we calculated the standard deviation between trials and the average correlation between answers given and correct answers at test time for different LLMs. To more clearly ascertain the difference between different LLMs in ophthalmic tests, the scores and correlation mean with their respective variances were counted and calculated.

To quantify the answer confidence of the LLM, we counted the number of correct answers for each question across all tests. For example, if each LLM answered the same question correctly five times, the percentage of questions for which all five answers were correct would increase by 1% (since there are 100 questions). In addition, the test results were compared with the distribution that would have occurred if the examinee had guessed randomly. For random guessing, the expected number of correct answers across five trials averaged approximately $0.25 \times 5 = 1.25$, and the multiple-choice questions all had four choices. Using this value, it is possible to estimate the number of times the correct answer occurs for each question based on the resulting Poisson distribution.

Finally, the cumulative scores calculated by ChatGPT (GPT-3.5 and GPT-4) and PaLM2 were compared with human scores.

## 3. Results

### 3.1 Comparison of LLM scores with human scores

Raw scores and average test scores are shown in Figures 1 and 2, respectively, where the LLM average test score is the mean of the scores of five tests, and the human group average test score is the mean of the scores of each person in the group.

In the original scores of LLM, it can be seen that different LLMs have different performances in terms of total uncertainty and the number of correct answers to questions, and GPT-4 covers a wider range of correct answers to questions (Figure 1). The average test scores revealed that each LLM outperformed undergraduates in general, with GPT-3.5 and PaLM2 being slightly below the master's level, while GPT-4 showed a level comparable to that of attending physicians (Figure 2).

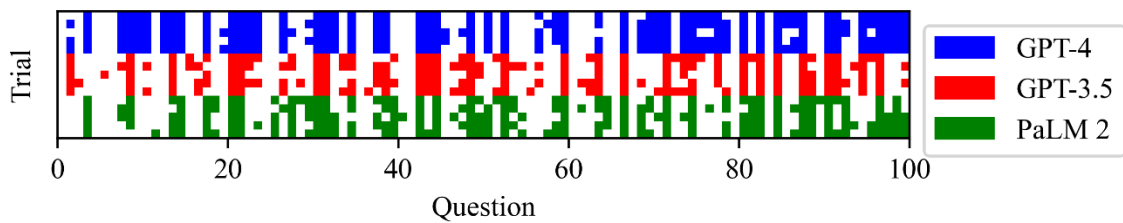

Figure 1. Heatmap Comparison of Answer Score Distributions for Different Large Language Models.

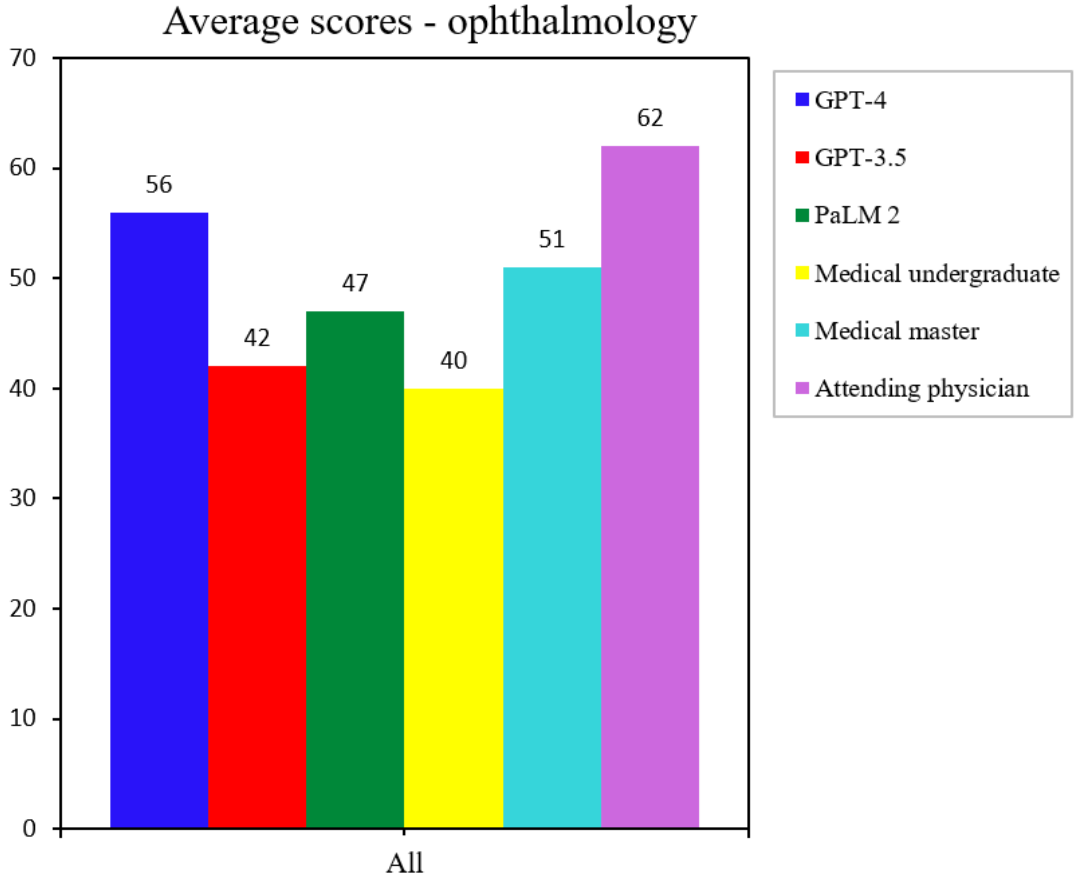

Figure 2. Comparison of Human Performance Scores on Ophthalmology Questions for Different Large Language Models across Various Educational Levels.

### 3.2 Comparison of the stability of LLM responses

The standard deviation and mean correlation across trials are shown in Figure 3. All three LLMs showed a high degree of agreement on scores and answers, with minimal standard deviation of scores (Figure 3A). GPT-4 showed a higher mean correlation (mean as high as 0.83) and a clear difference from GPT-3.5 (0.59) and PaLM2 (0.61), with PaLM2 showing a slightly higher mean correlation than GPT-3.5 but not a clear difference (Figure 3B).

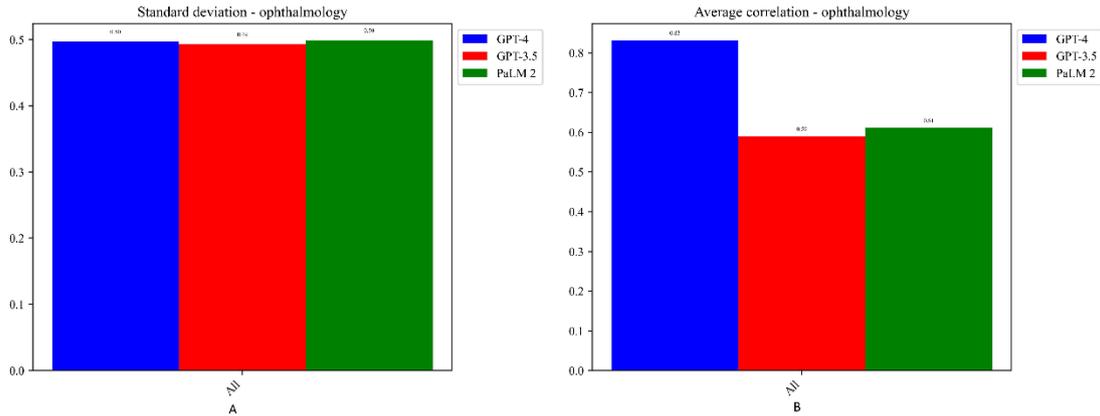

Figure 3. Analysis of Standard Error Distributions Related to Question Answer Accuracy and Answer Consistency for Different Large Language Models.

### 3.3 Comparison of self-confidence in LLM answers

From the results shown in Figure 4, the three LLMs bear no resemblance to random guessing, indicating that LLMs tend to present either confidence or confusion, with little possibility of randomly guessing the answer. GPT-3.5 (Figure 4A) showed 24% accuracy and 34% error. GPT-4 (Figure 4B) showed a high confidence on the test, giving 45% correct answers but still having a 37% probability of confusion. The levels of PaLM2 (Figure 4C) lied between that of GPT-3.5 and GPT-4.

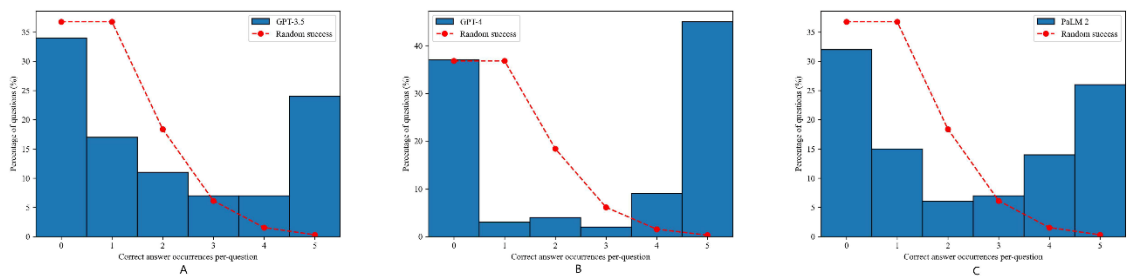

Figure 4. Comparison of self-confidence in LLM answers.

### 4. Discussion

Due to the increased popularity of electronic devices in modern life, the incidence of eye diseases not only shows an upward trend, but also shows a trend of younger age, which brings great pressure and burden to ophthalmology clinical work [26]. As ophthalmic professionals, it is essential to have a solid knowledge of ophthalmology to ensure safe and reliable treatment for patients. Furthermore, due to the uneven distribution of ophthalmologists in different regions, including urban and rural areas, there is significant disparity in healthcare access [27]. Effectively improving the ophthalmic diagnosis and treatment capabilities at the grassroots level can greatly enhance the prognosis of patients with eye diseases. Previous studies have compared the accuracy of treatment recommendations provided by a group of ophthalmology experts with those generated by AI. The consistency between the expert group and the AI group in treatment recommendations reached an accuracy of 61%. On the other hand, there was no significant difference between the expert group and the AI group in generating biased information[28]. These research findings demonstrate that LLMs have the potential for widespread and valuable applications in offering appropriate treatment recommendations for numerous eye disease patients. Google Research has leveraged fine-tuning with instruction prompts to develop its Path Language Model (PaLM) for the medical field, resulting in Med-PaLM, which achieved an accuracy rate of 85% in U.S. medical licensure exams[29]. This research further underscores the applicability of LLMs in the field of medicine.

This study aimed to evaluate the performance of LLM in a highly specialized subject as ophthalmology by creating 100 single-choice questions, and to compare the results with different levels of ophthalmology professionals in order to explore the reliability of LLM in medical education and clinical decision making. The 100 questions used for the test are listed in Appendix Section A. The results showed that the overall performance of the LLMs varied, with all three LLMs performing better than undergraduate medical students in general, medical masters performing slightly better than GPT-3.5 and PaLM2, and most surprisingly, GPT-4 performing almost on par with attending physicians. Medical undergraduates have a wide range of hunting in the medical field and contact with more basic and popular knowledge, which is equivalent to a non-entry level in the field of ophthalmology. However, the training data of LLM almost covers most of the knowledge learned by undergraduates. Thus, LLM easily exceeds the answer level of medical undergraduates. For medical masters, although they are not as good as attending doctors in ophthalmology knowledge reserve and clinical experience, they have also experienced systematic training and have a certain level of achievement in ophthalmology. Moreover, their logical reasoning ability allows them to get the answer that is unlikely to be wrong through reasoning and guessing, even if they do not know the correct answer, thus exhibiting a higher response level than GPT-3.5 and PaLM2. Attending doctors with a certain level of clinical work experience must have solid professional knowledge and clinical skills due to the nature of their work. GPT-4 showed a comparable response level to that of attending doctors, which is sufficient to prove that GPT-4 also performs very well in such a highly specialized topic as ophthalmology. Furthermore, in the evaluation of the American Academy of Ophthalmology Basic and Clinical Science Course Self-Assessment Program mock exam scores, even though GPT-4 received general cognitive training and did not undergo specific medical domain data training, its accuracy in text-based practice questions was significantly higher than that of ophthalmology residents and practicing ophthalmologists[30]. This is consistent with our research findings, indicating that general artificial intelligence has the potential for valuable applications in the field of ophthalmology.

Although LLM was generally better than medical undergraduates, GPT-4 showed a higher correlation than the other two LLMs, indicating that GPT-4 was more likely to choose the same answer for the same question in different tests. Compared with humans, GPT-4 also has an advantage in response correlation, because if the same person is repeatedly tested in the same way as the LLM, the person may also be confused on some questions and unable to choose the same answer each time, showing a certain degree of confusion. LLMs differ from random guesses in terms of answer confidence, as they always present as either confident or confused. At this point, GPT-4 again stood out among the three. Although GPT-4 also showed a certain degree of confusion, with a 37% probability of confusion, it outperformed GPT-3.5 and PaLM2, with an accuracy rate of 45%.

The above results show that GPT-4 performs well in all aspects, can achieve a similar answer level to that of attending doctors, and can always choose the same answer for the same question in different tests, showing a high degree of confidence in answering questions. Despite this, GPT-4 cannot completely replace the attending doctor due to several reasons. First, GPT-4 is absolutely confident when answering a question if it gives a correct answer, but when it gives a wrong answer, it will always confidently choose the same wrong answer and believe that it is correct. The human attending physician thinks on the basis of his own experience on highly specialized subjects, knows when to guess and how to guess wisely, and even if he is not sure about the right answer to some questions, he will be able to reason to arrive at an answer that is unlikely to be wrong. Secondly, even if the attending doctors have the same educational experience and professional background, their personal ability and knowledge reserve will be very different, and they have more in-depth research in some specific fields. Finally, answering questions is not completely equivalent to the daily work of a clinician. Clinical work is complex and tedious, and solid professional knowledge is only one element. It can be seen that only using the problems solving ability of LLM to evaluate its performance in clinical work may produce certain deviations.

Our study shows that LLM, especially GPT-4, holds great promise in ophthalmology, but there are many highly specialized areas in medicine that need to be evaluated. With the continuous upgrading of LLM, it is inevitable to use LLM to complete more high-end and complex tasks. Therefore, human beings should shift the focus of LLM from acquiring simple basic knowledge to more highly specialized fields. Furthermore, systems based on LLMs are not designed to replace ophthalmologists. Instead, they can enhance the work of ophthalmologists to some extent. For example, patients with eye diseases can consult an LLM system for preliminary information about common eye conditions before their eye clinic appointments. After the consultation, the model can further summarize and provide specific personalized visit summaries and follow-up recommendations. This approach allows for personalized patient education while saving valuable time for ophthalmologists and can be applied to more complex and challenging cases.

Of course, it is undeniable that in a highly specialized field such as ophthalmology, LLM represented by GPT-4 performs better. Our study suggests that with further modifications, GPT-4 may bring unexpected benefits in medical education and clinical decision making in the near future.

**4.1 Application of LLM in ophthalmology**

The powerful function of LLM makes it widely used in ophthalmology, and ChatGPT has been tested in ophthalmology. Bernstein et al.[31] conducted a study evaluating the quality of ophthalmic recommendations generated by LLM chatbots compared to recommendations written by ophthalmologists and found that the answers generated by chatbots did not differ from human answers in terms of guidance, safety, and reliability. LLM can be used as a platform to provide useful opinions across language barriers and meet the counseling needs of patients when they have eye problems[32]. Most online consultations are performed for common and frequently occurring diseases of ophthalmology. LLM can also give instructive opinions when faced with rare eye diseases that are not representative in the training dataset. Hu et al. evaluated the ability of ChatGPT-4 to diagnose rare eye diseases and found that it has the potential to be used as an auxiliary tool by primary ophthalmologists to diagnose rare eye diseases[33]. Meanwhile, ChatGPT can also be used to write surgical records, discharge summaries, and other documents[34, 35]. ChatGPT can record surgical procedures[36], specific drugs[37], follow-up instructions[38], and other information in the right place of the document with appropriate prompts. Doctors only need to make fine adjustments to produce medical documents that can be archived, which can undoubtedly greatly improve the efficiency of doctors. Even in research and scientific writing, LLM can guide researchers through different stages of scientific research, from idea generation to completion of manuscript writing[39].

**4.2 The future of LLM in ophthalmology**

Although LLM has been widely used in the field of ophthalmology, which has freed the hands of doctors to a certain extent, its biggest limitation is that it cannot process images. Ophthalmology is a field that relies heavily on visual examination and imaging for diagnosis, treatment, and follow-up. This requires us to extend LLM by incorporating LLM with some converter models that can handle multiple types of data[40]. For example, the contrastive language-image pre-training model can collect image information and generate text description, and LLM can use this description to answer questions[41]. In other words, LLM acquires information by adding an image modality to realize a multi-modality input. Multimodal models will be the future of LLM.

Furthermore, the specific analysis methods and learning processes of GPT in generating responses to user input are unknown, similar to machine learning methods, and they lack clear interpretability, making it somewhat challenging for clinical professionals and users to fully embrace them. We need to further enhance the interpretability of these "black-box models." Lastly, in terms of legal and regulatory considerations, AI products like ChatGPT, if intended for use in guiding clinical practice, must comply with relevant regulations set by the National Medical Products Administration to ensure the safe, reliable, and controllable development of AI technology in healthcare and address issues of responsibility and accountability in medical AI.

Moreover, the future of LLM is not to replace the role of medical experts, but to enhance professional judgment[42-44]. The success[45-48] of foundational models extend beyond NLP[15, 43, 49, 50]. Future developments should focus on improving the ability of LLMs to interpret complex medical terms and obtain image information[48], thereby improving their reliability in medical education and clinical decision making[10].

## 5、Conclusion

Our study shows that LLM represented by GPT-4 performs better in highly specialized fields such as ophthalmology, with high answer accuracy, stability, and confidence. With the continuous improvement of LLM, LLM will bring unexpected benefits to people in medical education and clinical decision making.

**Author contributions**

**Conflict of interest statement**

**Acknowledgments**

**Funding details**